\PassOptionsToPackage{dvipsnames,table}{xcolor}
\documentclass{article}

\usepackage[preprint]{neurips}

\usepackage[utf8]{inputenc} % allow utf-8 input
\usepackage[T1]{fontenc}    % use 8-bit T1 fonts
\usepackage{url}            % simple URL typesetting
\usepackage{booktabs}       % professional-quality tables
\usepackage{amsfonts}       % blackboard math symbols
\usepackage{nicefrac}       % compact symbols for 1/2, etc.
\usepackage{microtype}      % microtypography
\usepackage{xcolor}         % colors

\usepackage{multirow, booktabs}
\usepackage[usestackEOL]{stackengine}
\usepackage[dvipsnames]{xcolor}

\usepackage{graphicx}
\usepackage{booktabs}

\usepackage[accsupp]{axessibility}  % Improves PDF readability for those with disabilities.

\usepackage{graphicx}
\usepackage{tikz}
\usepackage{comment}
\usepackage{amsmath,amssymb} % define this before the line numbering.}
\usepackage{bbm}
\usepackage[colorlinks]{hyperref}
\usepackage{subfig}
\usepackage{bm}
\usepackage{ulem} 

\usepackage{algorithm}
\usepackage{algorithmicx}

\usepackage{mdframed}
\usepackage{orcidlink}

\usepackage{amstext}
\usepackage{amssymb}
\usepackage{booktabs}

\usepackage{tikz}
\usepackage{comment}
\usepackage{amsmath,amssymb} % define this before the line numbering.}
\usepackage{bbm}
\usepackage{booktabs}
\usepackage{graphicx}
\usepackage{bm}
\usepackage{multirow}
\usepackage{cases}
\usepackage{color}
\usepackage{ulem}
\usepackage[table]{xcolor} 
\definecolor{dino}{RGB}{249,231,227}

\usepackage{url}
\usepackage{epsfig}
\usepackage{amsthm}
\usepackage{multirow}
\usepackage{soul}
\usepackage{footnote}
\usepackage{tablefootnote}
\usepackage{caption}
\usepackage{subcaption}
\usepackage{mathtools}

\usepackage{url}
\usepackage{multirow}
\usepackage{booktabs}
\usepackage{pifont}
\usepackage{xcolor}
\usepackage{graphicx}
\usepackage{amsmath}
\usepackage{mathtools}
\usepackage{algpseudocode}
\usepackage{listings}
\usepackage{wrapfig}
\usepackage{subcaption}
\usepackage{xcolor,colortbl}

\definecolor{top1}{RGB}{255,179,179}
\definecolor{top2}{RGB}{255,217,179}
\definecolor{top3}{RGB}{255,255,179}

\usepackage{tikz}
\usepackage{comment}
\usepackage{amsmath,amssymb} % define this before the line numbering.}
\usepackage{bbm}
\usepackage{booktabs}
\usepackage[colorlinks]{hyperref}
\usepackage{graphicx}
\usepackage{bm}
\usepackage{multirow}
\usepackage{pifont}
\usepackage{xcolor,pifont}
\newcommand*\colourcheck[1]{%
  \expandafter\newcommand\csname #1check\endcsname{\textcolor{#1}{\ding{52}}}%
}
\colourcheck{blue}
\colourcheck{green}
\colourcheck{red}
\usepackage{makecell}

\newcommand{\tablestyle}[2]{\setlength{\tabcolsep}{#1}\renewcommand{\arraystretch}{#2}\centering\scriptsize}

\title{
GigaVideo-1: Advancing Video Generation via Automatic Feedback with 4 GPU-Hours Fine-Tuning
}

\author{
Xiaoyi Bao$^{1,2,3}$\footnotemark[1]\quad
Jindi Lv$^{1,4}$\footnotemark[1]\quad 
Xiaofeng Wang$^{1}$\quad
Zheng Zhu$^{1}$\footnotemark[2]\quad 
\\
\textbf{Xinze Chen}$^{1}$\quad
\textbf{YuKun Zhou}$^{1}$\quad
\textbf{Jiancheng Lv}$^{4}$\quad
\textbf{Xingang Wang}$^{2}$\footnotemark[2]\quad
\textbf{Guan Huang}$^{1}$\quad
\\
$^{1}$GigaAI\;
$^{2}$Institute of Automation, Chinese Academy of Sciences\;
\\
$^{3}$School of Artificial Intelligence, University of Chinese Academy of Sciences\;
\\
$^{4}$School of Computer Science, Sichuan University\;
\\
Project page: \href{https://gigavideo-1.github.io}{https://gigavideo-1.github.io}
}

\begin{document}

\maketitle
\footnotetext[1]{Equal contribution.}\footnotetext[2]{Corresponding Authors.}
\begin{abstract}
Recent progress in diffusion models has greatly enhanced video generation quality, yet these models still require fine-tuning to improve specific dimensions like instance preservation, motion rationality, composition, and physical plausibility. Existing fine-tuning approaches often rely on human annotations and large-scale computational resources, limiting their practicality. In this work, we propose GigaVideo-1, an efficient fine-tuning framework that advances video generation without additional human supervision. Rather than injecting large volumes of high-quality data from external sources, GigaVideo-1 unlocks the latent potential of pre-trained video diffusion models through automatic feedback. Specifically, we focus on two key aspects of the fine-tuning process: data and optimization. To improve fine-tuning data, we design a prompt-driven data engine that constructs diverse, weakness-oriented training samples. On the optimization side, we introduce a reward-guided training strategy, which adaptively weights samples using feedback from pre-trained vision-language models with a realism constraint. We evaluate GigaVideo-1 on the VBench-2.0 benchmark using Wan2.1 as the baseline across 17 evaluation dimensions. Experiments show that GigaVideo-1 consistently improves performance on almost all the dimensions with an average gain of \textasciitilde4\% using only 4 GPU-hours. Requiring no manual annotations and minimal real data, GigaVideo-1 demonstrates both effectiveness and efficiency. Code, model, and data will be publicly available.

\end{abstract}

\section{Introduction}
\label{submission}
In recent years, text-to-video (T2V) generation~\cite{gupta2024photorealistic,he2024venhancer,wang2023modelscope,wang2023videocomposer,yang2024cogvideox} powered by diffusion models has made substantial progress. These models can synthesize high-quality, stylistically diverse videos from natural language prompts and have been increasingly adopted in user-facing creative applications. Their success is largely driven by large-scale training data and extensive computational resources. However, despite strong performance in surface-level attributes such as per-frame aesthetics, temporal smoothness, and basic prompt adherence, these models still struggle with deeper aspects of video quality. Deficiencies in physical plausibility, object permanence, motion realism, and scene dynamics remain common, often diminishing the perceived coherence and credibility of generated content.

To mitigate these limitations, recent work has introduced fine-tuning pipelines aimed at improving generation quality along specific dimensions. These approaches generally fall into two categories. The first follows a supervised fine-tuning (SFT) paradigm, such as in Wan2.1~\cite{wan2025wan} and HunyuanVideo~\cite{kong2024hunyuanvideo}, where large-scale, high-quality datasets are used to continue training after pre-training. While effective, they demand extensive curated data and high computational cost. The second category, exemplified by methods like VideoAlign~\cite{liu2025improving} and LiFT~\cite{wang2024lift}, adopts human-in-the-loop reinforcement learning~\cite{liu2024videodpo,xu2024visionreward}. These methods require manually annotated data with preference scores and the training of reward models, making them difficult to scale and resource-intensive.

To address these challenges, we introduce GigaVideo-1, a lightweight and data-efficient fine-tuning framework designed to enhance pre-trained video diffusion models without requiring human supervision. Unlike prior methods that inject large amounts of high-quality external data during fine-tuning, GigaVideo-1 instead exploits the latent potential of pre-trained models through automatic feedback mechanisms. This enables the model to self-improve based on its own behavior, dramatically reducing the need for curated datasets.

Our method targets two key aspects of the fine-tuning process: training data and optimization. To improve data quality, we propose a prompt-driven data engine that generates diverse training samples specifically oriented toward challenging or underrepresented generation attributes. For optimization, we introduce a reward-guided training strategy. This method leverages feedback from frozen vision-language models to adaptively weight training samples and applies regularization to maintain alignment with real-world data distributions.

Together, these components allow GigaVideo-1 to improve model performance across critical dimensions with only a small number of synthetic samples and minimal computation. Using 4 GPU-hours, our approach demonstrates that high-quality fine-tuning of T2V models can be achieved efficiently. Without human annotation or large-scale external data, GigaVideo-1 marks a promising step toward scalable, automatic refinement for video generation.

Our contribution can be summarized as follows:
\begin{itemize}
    \item We propose GigaVideo-1, a lightweight fine-tuning pipeline for video generation. It improves pre-trained video diffusion models using automatic feedback without human supervision or large-scale data.

    \item From the aspect of fine-tuning data, we propose a prompt-driven data engine. It constructs diverse training samples by generating and expanding prompts specifically oriented toward model weaknesses.

    \item From the aspect of fine-tuning optimization, we propose a reward-guided training strategy. It adaptively weights training samples using feedback from pre-trained vision-language models and maintains distributional alignment with real data.

    \item Extensive experiments demonstrate the effectiveness and efficiency of our proposed method. Using minimum computation cost and external data, GigaVideo-1 realizes \textasciitilde4\% performance gain over the baseline Wan2.1~\cite{wan2025wan}, superior in almost all dimensions of the VBench-2.0 benchmark.
\end{itemize}

% Placeholder for Teaser Figure
\begin{figure*}[t]
    \centering
    \includegraphics[width=0.99\linewidth]{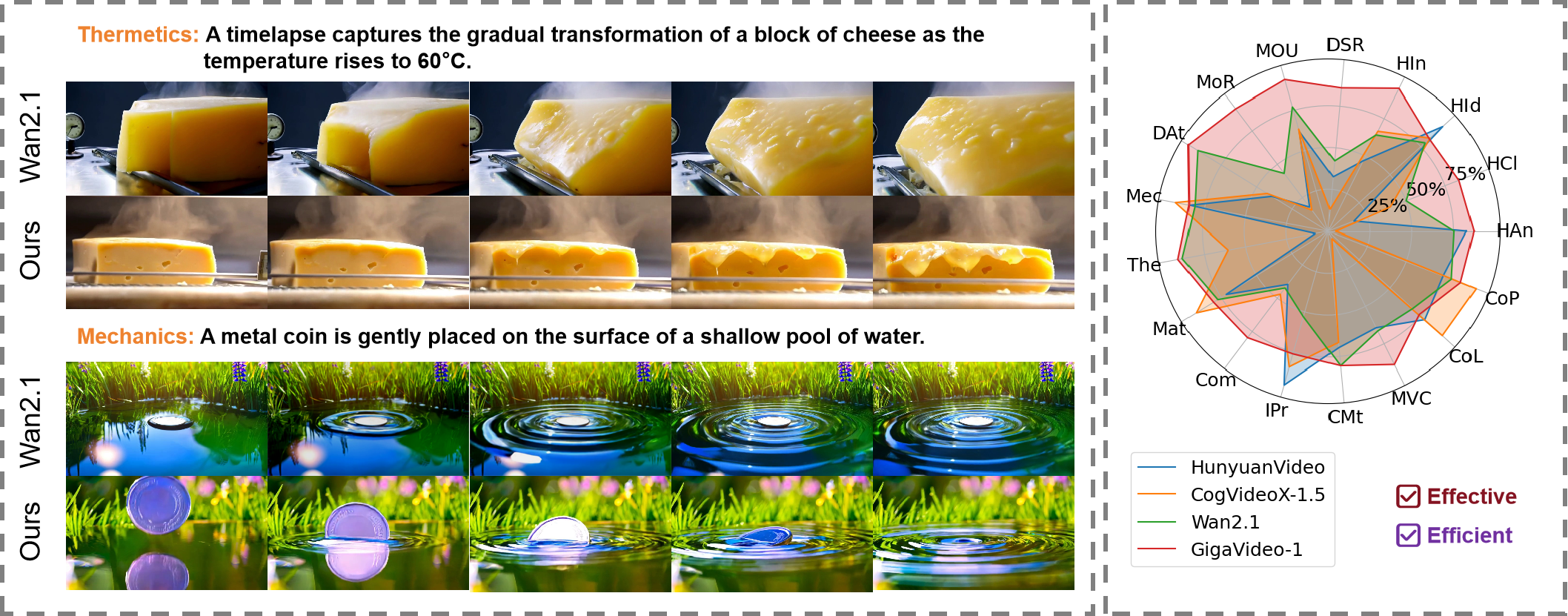} % Replace with your actual teaser figure
    %\fbox{\rule{0pt}{2in} \rule{0.9\linewidth}{0pt}} % Placeholder box
    % \vspace{-1em}
    \caption{
\textbf{Visualization of GigaVideo-1 performance}. The left figure compares videos generated by the baseline Wan2.1~\cite{wan2025wan} and our GigaVideo-1 across two different dimensions. The right figure provides the performance of GigaVideo-1 and other state-of-the-art T2V models on VBench-2.0. With only ~4 GPU-hours of training, GigaVideo-1 achieves notable improvements over the baseline, demonstrating both effectiveness and efficiency.}
%微调前后的可视化对比+微调前后分数多边形对比图/和human-in-the-loop方法的框架对比+效率} % Updated caption
    \label{fig:teaser} % Renamed label
    \vspace{-5pt}
\end{figure*}

\section{Related Work}

\subsection{Text-to-Video Generation Models}
Video generation, particularly diffusion-based models~\cite{blattmann2023stable,chen2023videocrafter1,harvey2022flexible,ma2024latte,zheng2024open,mei2023vidm}, has witnessed rapid progress in recent years. Among them, text-to-video~\cite{singer2022make,zhang2024show,zhou2024allegro} generation has found broad applications in personalized content creation, enabling controllable video synthesis based on user-provided prompts. A growing amount of research has demonstrated that scaling up both the data volume and model size leads to improved performance. As a result, numerous large-scale models have emerged, achieving strong results on general video generation tasks but often requiring extensive training data and computational resources. On the model structure side, researchers have also explored various innovations to improve model expressiveness and efficiency. For example, some frameworks replace the standard U-Net in diffusion models with transformer-based backbones DiT~\cite{peebles2023scalable} to enhance representation capacity. Some models, like Hunyuan~\cite{kong2024hunyuanvideo}, introduce 3D-VAE to better capture spatiotemporal dynamics. Another group of approaches (like SVD~\cite{blattmann2023stable} and Wan~\cite{wan2025wan}) uses improved modeling of the diffusion process, applying EDM~\cite{karras2022elucidating} and flow-matching techniques~\cite{esser2024scaling} to accelerate diffusion inference and improve modeling efficiency, respectively. While these models perform well on general prompts, their outputs often fail to exhibit reasonable behavior in more structured or physics-intensive scenarios. For example, generated videos may violate basic physical principles such as object permanence or causality. New benchmarks such as VBench-2.0~\cite{zheng2025vbench}, Worldmodelbench~\cite{li2025worldmodelbench} evaluate the effectiveness of the existing models in the above aspects. The results show that there is significant room for improvement in the behavior of these dimensions.

\subsection{Video Generation Fine-Tuning}
Fine-tuning has emerged as an essential step for improving specific aspects of video generation, such as aesthetics and text alignment. Current methods can be broadly categorized into supervised fine-tuning (SFT) and reinforcement learning (RL)-based alignment, each with distinct advantages and limitations. Supervised fine-tuning (SFT) methods, such as Wan2.1~\cite{wan2025wan} and HunyuanVideo~\cite{kong2024hunyuanvideo}, extend model training using large-scale, high-quality datasets. These methods are effective at improving general-purpose generation, but they require significant computational resources and manual data curation, making them costly and inefficient. 

The other fine-tuning strategy is reinforcement learning-based, aiming to align generation with human preferences using feedback-driven supervision. Among them, reward-weighted regression (RWR)~\cite{peng1910advantage,lee2023aligning,furuta2024improving,liu2025improving,he2022latent} assigns scalar rewards to generated samples and reweights their training loss. Although computationally efficient, methods like LiFT~\cite{wang2024lift} and FlowRWR~\cite{liu2025improving} still rely on human-annotated data to train reward models. Another widely adopted class, direct preference optimization (DPO)~\cite{rafailov2023direct,wallace2024diffusion,dong2023raft,yang2024using,liang2024step}, learns from pairwise human preference comparisons or ranked samples to guide model adaptation. Representative works include FlowDPO~\cite{deng2024flow} and GAPO~\cite{gu2025gapo}. Despite improved alignment with subjective quality, these methods are also constrained by the need for curated human preference data. Other RL methods, such as PPO~\cite{schulman2017proximal,black2023training,fan2023reinforcement} and online backpropagation, are less common due to their high sampling cost and instability in high-dimensional video spaces.

While the above existing fine-tuning methods have demonstrated impressive improvements in generation quality, they either depend on massive training data with huge computation cost, or strongly rely on costly human-in-the-loop annotations. 
\section{Method}

% 3.1 Method Overview
\subsection{Framework Overview} % Changed heading from \noindent\textbf
We propose a lightweight and automatic fine-tuning pipeline for video generation, designed to enhance model performance efficiently using minimal data and compute. As a starting point, Sec.~\ref{sec:prelimi} revisits the standard training formulation for text-to-video models, which centers on learning from real video–text pairs under a fixed objective.
Building on this foundation, our pipeline introduces two key modifications, as illustrated in Fig.\ref{fig:data_pipeline}: data construction and optimization strategy. Sec.~\ref{sec:data} presents a prompt-driven data engine that diversifies prompt coverage and calibrates the training data distribution using synthetic video generation. Sec.~\ref{sec:reward} describes our reward-guided optimization strategy, which leverages automatic feedback to adaptively weight training samples while preserving alignment with real-world distributions.

Together, these components form a modular and efficient framework for fine-tuning pre-trained video generation models without human annotations.

\subsection{Preliminary}\label{sec:prelimi} % Changed heading from \noindent\textbf

Recent advances in video generation~\cite{esser2023structure,brooks2024video,chen2023videocrafter1,wang2023modelscope} have shifted from traditional denoising-based objectives toward more principled training formulations. Among them, flow-matching has emerged as an effective approach for training diffusion-based models. It frames generation as learning a continuous transformation from noise to data through ordinary differential equations (ODEs), offering both stability and theoretical alignment with maximum likelihood training.

In this setting, training involves linearly interpolating between a random noise sample $z_0 \sim \mathcal{N}(0, I)$ and a clean video latent $z_1$. Following Rectified Flows (RFs)~\cite{esser2024scaling}, the model is trained to predict the ground-truth velocity $v_{t}$ between these two endpoints. The intermediate latent $z_t$ and the ground-truth velocity $v_{t}$ are formed as:
\begin{equation}\label{eq:z_t}
    z_{t}=t z_{1}+(1-t) z_{0},     v_{t}=\frac{d z_{t}}{d t}=z_{1}-z_{0}.
\end{equation}
The learning objective minimizes the discrepancy between the predicted and actual velocity $v_{t}$, conditioned on the prompt embedding and timestep:
\begin{equation}\label{rf_loss}
\mathcal{L} = \mathbb{E}_{t,z_0, (z_1, p) \sim \mathcal{D}^{\mathrm{real}}}||u(z_t, p, t; \theta)-v_t||^2,
\end{equation}
where $p$ is the text-form prompt, $\theta$ is the model weights, and $u(z_t, p, t; \theta)$ denotes the output velocity predicted by the model. This formulation reveals two core factors that directly impact fine-tuning performance: 1). The training samples. The way we construct prompts and videos determines what types of visual behaviors the model will learn to refine. This makes it crucial to expose the model to failures it needs to correct. 2). The optimization strategy. The loss determines how each sample contributes to training. A static formulation treats all samples equally, while more informative or underperforming cases should be prioritized. 

These two perspectives form the basis of our fine-tuning pipeline. In the following sections, we introduce methods to re-design both components for effective, automatic fine-tuning of pre-trained video generation models.

% 3.2 data-driven
\subsection{Prompt-Driven Data Engine}\label{sec:data}
% Placeholder for Overview/Pipeline Figure
\begin{figure}[t]
    \centering
    \includegraphics[width=\linewidth]{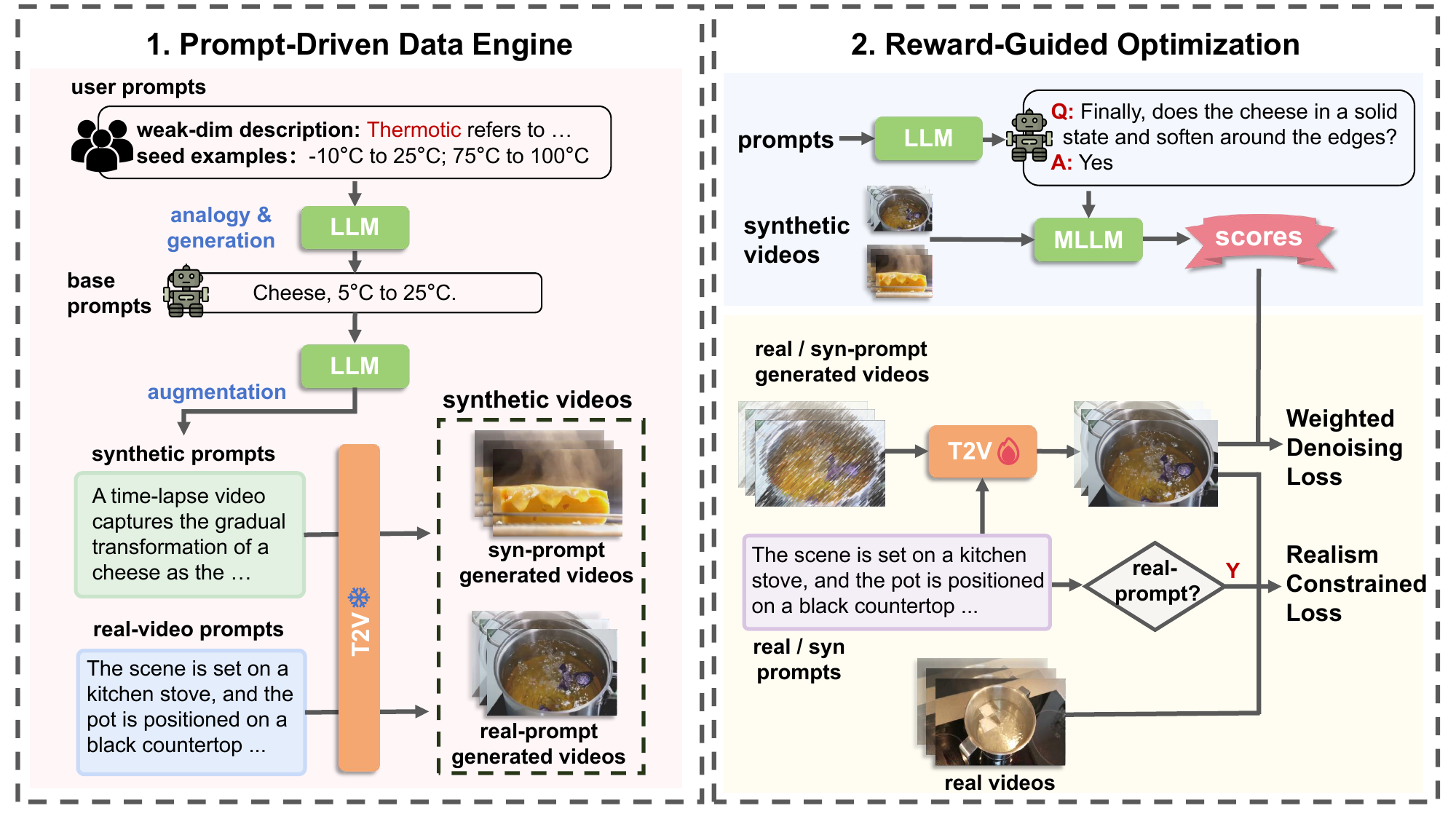} % Replace with your actual pipeline figure
    %\fbox{\rule{0pt}{2in} \rule{0.9\linewidth}{0pt}} % Placeholder box
    % \vspace{-1em}
\caption{
\textbf{GigaVideo-1 Training Pipeline.} Our pipeline consists of two components: prompt-driven data engine and reward-guided optimization. On the left, we generate synthetic prompts targeting weak dimensions using LLMs, and synthesize training videos via a pre-trained T2V model. These are combined with real-caption–based samples to balance diversity and realism. On the right, a frozen MLLM scores each video on dimension-specific criteria. These scores guide training via weighted denoising loss. For synthetic videos from real-world caption prompts, extra realism constraint is applied for distribution alignment. GigaVideo-1 enables efficient, automatic fine-tuning without manual labels or extra data collection.}
    \label{fig:data_pipeline} % Renamed label
    \vspace{-10pt}
\end{figure}

Traditional video generation methods typically rely on real-world video-caption pairs to supervise training.
While effective for general-purpose modeling, such data often reflects the natural distribution of events, scenes, and interactions. Elements like background clutter, occlusions, and multi-object dynamics are common. These factors dilute supervision when the goal is to improve specific weaknesses, such as unrealistic motion or poor physical consistency.

To address this challenge, we introduce a prompt-driven data engine designed to construct targeted and controlled training samples. Rather than relying on captions extracted from real videos, we aim to generate synthetic prompts that directly highlight the dimensions where pre-trained models perform poorly. This allows us to bypass the uncontrolled variability of real-world content and focus learning on specific failure modes.

We start by identifying specific failure modes in the pre-trained model (e.g., unrealistic motion, implausible interactions) and selecting a few representative prompt examples that expose these issues. Each weak dimension is defined with a short description and a handful of seed phrases. These are then fed into a large language model using few-shot prompting to generate a diverse set of base prompts that are concise, visually grounded, and explicitly focused on the target concept (e.g., “Camera zoom in, Forbidden City”). More details and examples of the constructed base prompts can be found in the supplementary material.

We further expand these base prompts into stylistic variants by modifying scene elements such as lighting, motion type, or character appearance. This augmentation increases data diversity without deviating from the intended training objective. These prompts are then used to condition a pre-trained text-to-video model to generate synthetic videos. Compared to real-world footage, these synthetic samples are more focused and better aligned with the target dimensions.

This pipeline enables controlled data construction and effective fine-tuning without reliance on large-scale external human annotations. By eliminating irrelevant factors present in real-world data and amplifying targeted supervision signals, our method improves training efficiency and accelerates model adaptation on weak dimensions, even under limited computational budgets.

\subsection{Reward-Guided Optimization}\label{sec:reward}
In conventional video generation training, the weighting coefficient $\lambda$ is typically set as a fixed constant to control the scale of the loss. However, this static setting fails to reflect the varying quality and usefulness of individual training samples, particularly when focusing on improving specific aspects of generation.

To address this, we propose a reward-guided optimization that leverages automatic feedback from pre-trained vision-language models. To evaluate synthesized video–prompt pairs in a dimension-specific and interpretable manner, we leverage a large language model (LLM) to transform abstract quality dimensions into concrete, checkable criteria grounded in world knowledge. For each video, we first provide its associated prompt along with a target evaluation dimension (e.g., motion consistency or physical plausibility) to the LLM, which then generates a set of question–answer (QA) pairs. These QA pairs are typically yes/no format and represent expected outcomes if the video correctly exhibits the intended behavior, serving as explicit checks for the target dimension, as the example in Fig.~\ref{fig:data_pipeline}.

We then feed both the video and the generated QA pairs into a frozen multimodal large language model (MLLM), which internally assesses whether the video content aligns with the expectations posed by the questions. Rather than returning specific answers, the MLLM directly outputs a scalar score reflecting the degree of agreement between the video and the reference answers. Videos that more closely match the dimension-specific world knowledge encoded in the QA pairs receive higher scores. This enables scalable, fine-grained evaluation without requiring human annotation. For a small set of dimensions that require fine-grained visual perception, we employ specialized pre-trained vision models like CoTracker2~\cite{karaev2024cotracker}. Detailed descriptions of these modules can be found in the supplementary material.

The resulting scores serve as sample-level rewards, which we use to dynamically adjust the training weight for each sample. Formally, the reward-weighted training loss is:

\begin{equation}\label{rf_loss}
\mathcal{L}_{P_s}(\theta) = \mathbb{E}_{t,z_0, (z_1, x, p) \sim \mathcal{D}^{\mathrm{syn}}}\left[-r_\phi(x, p) \cdot  ||u(z_t, p, t; \theta)-v_t||^2\right],
\end{equation}

where $z_{1}$ is the clean latent of video $x$, $r_\phi(x, p)$ is the score given by reward model. This training pipeline allows the model to place greater emphasis on high-quality or informative samples during training, while deprioritizing noisy or less useful ones. Notably, this effective training process is realized automatically, without requiring any human-labeled data.

To prevent the model from overfitting to synthetic distributions or drifting away from realistic video properties, we introduce a realism constraint. For prompts derived from captions of real videos (as discussed in Sec.~\ref{sec:data}), we generate corresponding synthetic videos using the same pipeline and include both the original and synthetic pairs during training. Additionally, we apply a KL divergence loss to encourage the model's outputs to remain consistent with the distribution of real videos. 

With the predicted velocity $u(z_t, p, t; \theta)$, the predicted clean latent can be represented by $u(z_t, p, t; \theta)+z_0$, following the same formulation as Eq.~\ref{eq:z_t}. Formally, the training objective is:
\begin{equation}\label{rf_loss}
\begin{split}
\mathcal{L}_{P_r}(\theta) = \mathbb{E}_{t,z_0, (z_1,p) \sim \mathcal{D}^{\mathrm{real}},(\hat{x},\hat{z_1})\sim \mathcal{D}^{\mathrm{syn}}}\left[-r_\phi(\hat{x}, p) \cdot  ||u(z_t, p, t; \theta)-v_t||^2\right.\\\left.
+\lambda_{kl} \cdot D_{\text{KL}}((u(z_t, p, t; \theta)+z_0) \parallel z_{1})\right],
\end{split}
\end{equation}
where $z_{1}$ and $p$ represent the real video's latent and the video caption. $\hat{x}$ and $\hat{z_{1}}$ refer to a synthesized video generated from this caption and its latent embedding, respectively. Here, the first term encourages the model to learn from controlled, failure-mode-driven synthetic videos, while the second term preserves distributional alignment with real data. This dual-objective loss enables efficient fine-tuning to challenging dimensions without compromising its overall generation quality.

Overall, the total training loss consists of two parts, as shown in the following equation.
\begin{equation}
\begin{aligned}
\mathcal{L}(\theta) = &\ \lambda_{p_s} \cdot \mathcal{L}_{P_s}(\theta)
 + \lambda_{p_r} \cdot \mathcal{L}_{Pr}(\theta),
\end{aligned}
\end{equation}
where $\lambda_{p_s}$ and $\lambda_{p_r}$ are empirically set to 0.5 and 0.5.

\section{Experiments}
\subsection{Experimental Setups}
\begin{table*}[t!]
\centering
\begin{center}
\caption{\textbf{Experiments on single dimension of VBench-2.0.} We choose Wan2.1~\cite{wan2025wan} as our baseline. We compare the evaluation results of our fine-tuned GigaVideo-1 with the baseline and 4 other recent state-of-the-art video generation models across 17 VBench-2.0 dimensions. A higher score indicates better performance in the corresponding dimension.
}
\vspace{-2pt}
\resizebox{0.98\linewidth}{!}{
\renewcommand\arraystretch{1.5}
\begin{tabular}{c|c|c|c|c|c|c|c|c|c}
\toprule
\multirow{2}{*}{\textbf{Models}}   & \textbf{Human} & \textbf{Human} & 
\textbf{Human} & \multirow{2}{*}{\textbf{Composition}} & \multirow{2}{*}{\textbf{Mechanics}} & \multirow{2}{*}{\textbf{Material}} & \multirow{2}{*}{\textbf{Thermotics}} & \textbf{Multi-view}&\textbf{Dynamic Spatial}\\
& \textbf{Anatomy} & \textbf{Clothes} & \textbf{Identity} & & & & & \textbf{Consistency}&\textbf{Relationship}\\
\midrule% \Xhline{1pt}

HunyuanVideo~\cite{kong2024hunyuanvideo}        & 88.58 & 82.97 & 75.67 & 43.96 & 76.09 & 64.37 & 56.52 & 43.80&21.26 \\
CogVideoX-1.5~\cite{yang2024cogvideox}  & 59.72 & 87.18 & 69.51 & 44.70 & \textbf{80.80} & \textbf{83.19} & 67.13 & 21.79&19.32 \\ 
Sora~\cite{openaisora2024} & 86.45 & \textbf{98.15} & \textbf{78.57} & \textbf{53.65} & 62.22 & 64.94 & 43.36 & 58.22 & 19.81 \\ 
Kling 1.6~\cite{kuaishou2024kling}  & 86.99 & 91.75 & 71.95 & 43.89 & 65.55 & 68.00 & 59.46 & \textbf{64.38} & 20.77 \\
Wan2.1~\cite{wan2025wan}  & 85.87 & 89.00 & 67.02 & 44.23 & 74.42 & 69.64 & 72.66 & 44.60 &22.22\\
\rowcolor{dino}GigaVideo-1  &\textbf{90.18($\uparrow$4.31)}&95.10($\uparrow$6.1)&69.14($\uparrow$2.12)&47.96($\uparrow$3.73)&76.30($\uparrow$1.88)&72.32($\uparrow$2.68)&\textbf{73.19($\uparrow$0.53)}&52.87($\uparrow$8.27)&\textbf{26.57($\uparrow$4.35)} \\

\toprule
\multirow{2}{*}{\textbf{Models}} &   \textbf{Dynamic} & \textbf{Motion Order} & \textbf{Human} & \textbf{Complex} & \textbf{Complex} & \textbf{Camera} & \textbf{Motion} & \textbf{Instance}& \multirow{2}{*}{\textbf{Mean}}\\
&   \textbf{Attribute} & \textbf{Understanding} & \textbf{Interaction} & \textbf{Landscape} & \textbf{Plot} & \textbf{Motion} & \textbf{Rationality} & \textbf{Preservation}&\\
% \Xhline{1pt}
\midrule
HunyuanVideo~\cite{kong2024hunyuanvideo} & 22.71 & 26.60 & 67.67 & 19.56 & 10.11 & 33.95 & 34.48 & 73.79 &49.53\\ 
CogVideoX-1.5~\cite{yang2024cogvideox}   & 24.18 & 26.94 & 73.00 & \textbf{23.11} & \textbf{12.42} & 33.33 & 33.91 & 71.03 &48.90\\ 
Sora~\cite{openaisora2024} & 8.06 & 14.81 & 59.00 & 14.67 & 11.67 & 27.16 & 34.48 & 74.60 &47.64\\
Kling 1.6~\cite{kuaishou2024kling}  & 19.41 & 29.29 & 72.67 & 18.44 & 11.83 & \textbf{61.73} & 38.51 & \textbf{76.10}&52.98\\ 
Wan2.1~\cite{wan2025wan}  &46.15 &29.97&72.33&17.11&10.69&36.11&40.80& 63.26&52.12\\
\rowcolor{dino}GigaVideo-1  &\textbf{49.08($\uparrow$2.93)}&\textbf{33.67($\uparrow$3.70)}&\textbf{80.67($\uparrow$8.34)}&18.44($\uparrow$1.33)&11.33($\uparrow$0.64)&36.11($\uparrow$0)&\textbf{52.87($\uparrow$12.07)}&69.03($\uparrow$5.77)&\textbf{56.17($\uparrow$4.05)} \\
\bottomrule
\end{tabular}
}
\vspace{-4mm}
\label{tab:first}
\end{center}
\end{table*}

\noindent\textbf{Datasets and Implementation.} The training data consists of two main sources, as shown in Sec.~\ref{sec:data}. 1). The real-world dataset comprises \textasciitilde3.5k carefully curated video samples from the Koala~\cite{wang2024koala} dataset, each accompanied by an accurate and detailed caption and a corresponding synthetic video generated with this caption. 2). The synthetic dataset from synthetic prompts contains \textasciitilde9.5k videos generated by domain-specific prompts. We use different pre-trained T2V models for generation~\cite{kong2024hunyuanvideo,he2024videoscore,kuaishou2024kling,openaisora2024,wan2025wan}. More details about the training data, models, and settings can be found in the supplementary material.

\noindent\textbf{Dimension and Evaluations.} For selecting target weak dimensions, we follow VBench-2.0, a recently proposed benchmark that systematically investigates the limitations of current video generation models. VBench-2.0 decomposes overall video generation quality into 18 fine-grained, hierarchical dimensions, providing tailored prompts and evaluation protocols for each. Since the ``diversity'' dimension cannot be meaningfully assessed on a single synthesized video, we exclude it from our pipeline. The remaining 17 dimensions are used as our fine-tuning targets and serve as the benchmark for evaluating model performance.

\subsection{Evaluation Results}
\paragraph{Qualitative Comparison.} To evaluate the effectiveness of GigaVideo-1, we choose 17 dimensions from VBench-2.0 as instances of our specific dimensions. As detailed in Tab.~\ref{tab:first}, various aspects like human attribute, commonsense, and physical plausibility are considered.  We compare with our baseline Wan2.1~\cite{wan2025wan} and the other 4 state-of-the-art video generation models~\cite{kong2024hunyuanvideo,yang2024cogvideox,openaisora2024,kuaishou2024kling}. After applying our GigaVideo-1 fine-tuning pipeline, we observe substantial improvements over the baseline across most dimensions. Specifically, with only 4 GPU-hours of training for single dimension, our method achieves an average gain of approximately 4\% across the 17 dimensions. Notably, we observe an 8.27\% improvement in the Multi-view Consistency, 8.34\% improvement in the Human Interaction dimension and an increase of over 12\% in Motion Rationality. The average performance gain over Wan2.1~\cite{wan2025wan} is approximately 4\%. While the five advanced T2V generation models exhibit different strengths and weaknesses across various aspects, our GigaVideo-1 achieves the best overall performance after fine-tuning, outperforming the previous state-of-the-art on average.

Moreover, as shown in Tab.~\ref{tab:first}, the improvements brought by GigaVideo-1 vary across different evaluation dimensions. While our method delivers substantial gains in most cases, the improvements are relatively modest on certain dimensions. For example, there is no gain observed for Camera Motion. We hypothesize that this is due to the limited number of training samples that can be reliably labeled along this dimension. Specifically, for synthetic videos generated from real-world captions, dimensions such as Motion Rationality and Human Interaction are more frequently exhibited, enabling pre-trained MLLMs to provide consistent scores. In contrast, prompts that naturally evoke Camera Motion features are less common, as most videos tend to be static. Additionally, unlike other dimensions labeled by VLMs, Camera Motion scoring relies on CoTracker2, whose strict definitions of camera movement further reduce the number of training samples with valid feedback. We expect this limitation can be alleviated in future work by expanding the prompt pool within the prompt-driven data engine to better cover underrepresented dimensions.

\begin{figure}[t]
    \centering
    \includegraphics[width=0.95\linewidth]{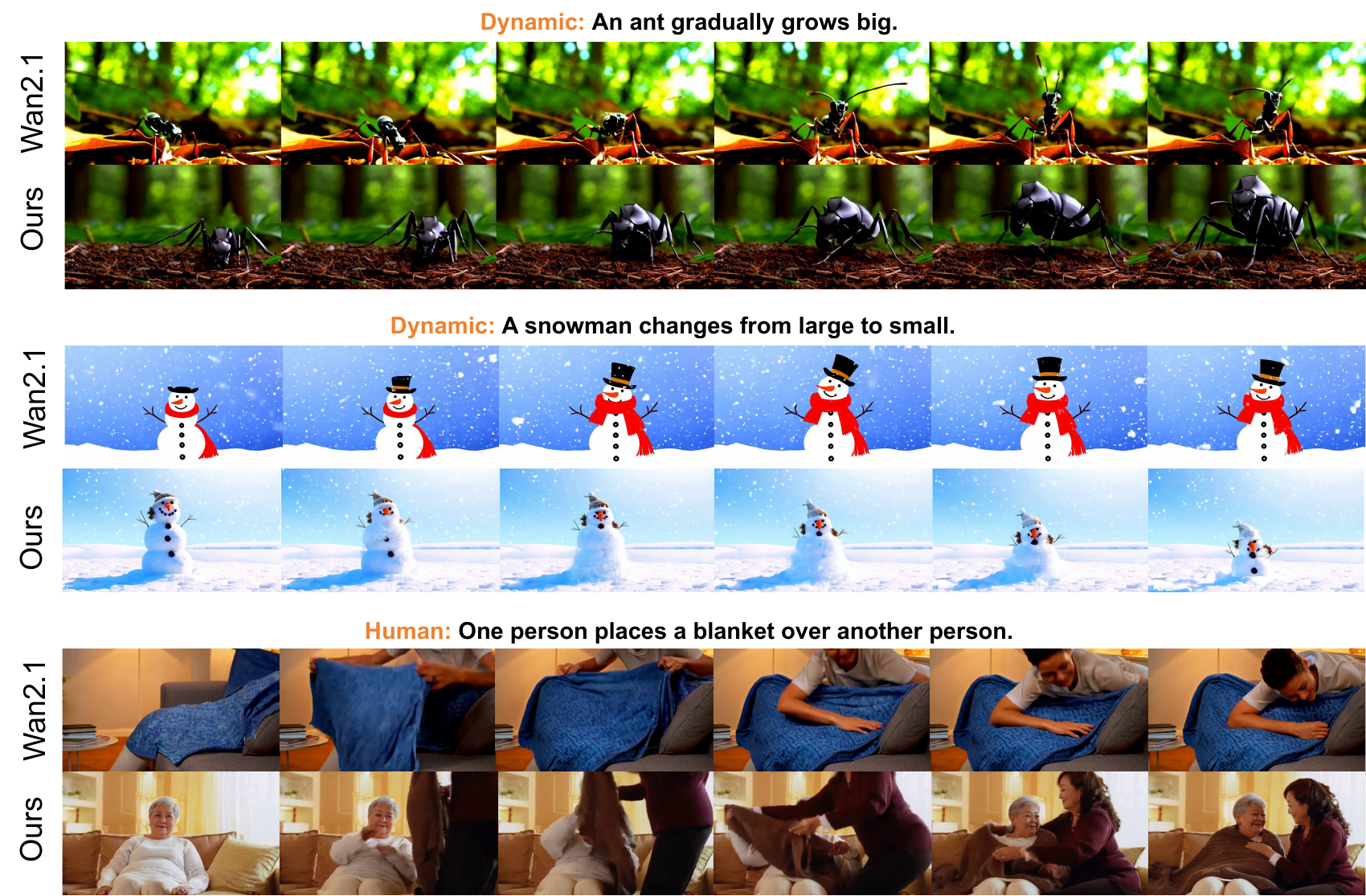} % Replace with your actual pipeline figure
    %\fbox{\rule{0pt}{2in} \rule{0.9\linewidth}{0pt}} % Placeholder box
    % \vspace{-1em}
    \caption{Visual comparisons of videos generated by GigaVideo-1 and Wan2.1~\cite{wan2025wan} across different real-world dimensions.}
    \label{fig:visualization} % Renamed label
    \vspace{-15pt}
\end{figure}

\paragraph{Quantitative Analysis.} 
We provide qualitative visualizations of generated videos in Fig.~\ref{fig:visualization}. As shown, GigaVideo-1 demonstrates clear improvements over the baseline, particularly on dimensions where the original model underperforms. The generated sequences, like the growth of an insect or the melting of a snowman, exhibit more realistic physical dynamics and align better with common-sense expectations. Scenes involving human interactions also appear more visually plausible.
We additionally conduct a user study to assess the quality preference of videos generated by Wan2.1 and GigaVideo-1 across the aforementioned evaluation dimensions. Details of the evaluation protocol and results are provided in the supplementary material.

\subsection{Ablation Studies}
We perform ablation studies to validate the key design choices of GigaVideo-1, specifically the impact of the prompt-driven data engine and the reward-guided optimization strategy.

\paragraph{Impact of Prompt-Driven Data Enhancement}
We investigate the effectiveness of the prompt-driven data engine, as summarized in Tab.~\ref{tab:data}. Specifically, we conduct an ablation study based on the source of training data, categorizing it into three types: (1) synthetic prompts with synthetic videos $P_sV_s$, (2) real captions with synthetic videos $P_rV_s$, and (3) real captions with real videos $P_rV_r$. From Tab.~\ref{tab:data}, we observe that using synthetic prompts ($P_sV_s$) leads to a significant improvement of 4\%, while training on synthetic videos generated from real captions ($P_rV_s$) yields a 6.67\% substantial performance gain. This indicates the effetiveness of these two data sources.

Interestingly, naively combining two synthetic videos ($P_sV_s + P_rV_s$) results in performance degradation to 73.66, compared to using $P_rV_s$ alone. We hypothesize that this is due to the introduction of unnatural concepts in synthetic prompts (e.g., “a person with a long tail”), which, while useful for enhancing dimensions like composition, may skew the distribution when not properly anchored by real-world data. Incorporating real videos ($P_rV_r$) alongside synthetic videos from real captions ($P_rV_s + P_rV_r$) improves performance further (to 79.33), suggesting that real videos offer valuable grounding and diversity. Finally, using all data sources together yields the highest performance, demonstrating that our full data pipeline benefits from the complementary strengths of each source. The addition of real videos mitigates the potential bias introduced by synthetic prompts, while the synthetic components enhance coverage of model weaknesses in a controlled manner.

\paragraph{Benefit of Reward-Guided Optimization}
We further investigate the impact of different reward-related optimization strategies, as presented in Tab.~\ref{tab:reward}. We begin by evaluating the effect of supervised fine-tuning (SFT) using all data generated from the prompt-driven data engine. As shown in Tab.~\ref{tab:reward}, applying SFT to the full dataset leads to a moderate improvement over the baseline, raising the average accuracy from 72.33\% to 75.67\%. However, the training process is relatively expensive, requiring 5.75 GPU-hours per epoch.

Compared to standard SFT, GigaVideo-1 achieves a notable accuracy gain of 5\% while reducing training time by more than 6$\times$. The efficiency arises from the process of reward-guided optimization. We filter out samples that cannot be reliably scored under the target dimension and samples that receive zero reward from gradient updates. To ensure the performance gain is not solely due to data filtering, we apply SFT to the same filtered dataset. While this reduces the training cost to a comparable level, the resulting accuracy remains at 75\%, confirming that our reward-guided optimization contributes meaningfully to both effectiveness and efficiency.

To further understand the role of reward signal acquisition and integration, we compare several optimization strategies. Implementation details and analysis of these strategies are provided in the supplementary material. Among them, our offline reweighting achieves the best high accuracy with minimal training overhead. These results demonstrate that our reward-guided optimization framework offers a highly efficient and effective alternative to conventional fine-tuning.

\begin{table}
    \centering
    \begin{minipage}{0.46\linewidth}
        \centering
          \caption{\textbf{Ablation of Data Engine}. ``$P_r$'' and ``$P_s$'' refer to using real video captions as prompts and generating metric-targeted prompts, respectively. ``$V_r$'' and ``$V_s$'' is using real and synthetic videos for reward labeling, respectively.}
             \tablestyle{14pt}{1.4}
              \begin{tabular}{ccc|c}
                \toprule
                $P_sV_s$ & $P_rV_s$ & $P_rV_r$ & Acc $\uparrow$ \\
                \midrule
                 & & &72.33  \\%(wanbaseline)
                \checkmark & & & 76.33 \\%(1kreweight)
                 & \checkmark& &79.00  \\%(3k5wokl)
                \checkmark& \checkmark&  &73.66  \\%(4k5wokl)
                &\checkmark  &\checkmark  &79.33  \\%(3k5wkl)
                \checkmark & \checkmark & \checkmark &80.67 \\%(4k5wkl)
                \bottomrule
              \end{tabular}
        \vspace{-8pt}
        \label{tab:data}
    \end{minipage} 
    \hfill
    \begin{minipage}{0.51\linewidth}
        \centering
            \caption{\textbf{Ablation of Reward Strategy}. ``SFT'' means vanilla fine-tuning, while ``filtered'' means only using samples with a positive score. ``Reweight / Backprop'' means using the reward scores as loss weight or direct loss for backpropagation. ``Online / Offline'' subscripts refer to different scoring stages of reward models.}
             \tablestyle{12pt}{1.2}
              \begin{tabular}{l|c|c}
                \toprule
                Method & Time(4GPU) $\downarrow$ & Acc $\uparrow$ \\
                \midrule
                Wan & - &  72.33\\
                Wan+$SFT$ & 5.75h/epoch & 75.67 \\%4k5sft
                Wan+$SFT (filtered)$ & 0.90h/epoch & 75.00 \\
                Wan+$Reweight_{online}$ &7.74h/epoch&79.33 \\
                Wan+$Backprop_{online}$ &7.50h/epoch&79.00  \\
                Wan+$Reweight_{offline}$ &0.90h/epoch&80.67 \\
                \bottomrule
              \end{tabular}
         \vspace{-8pt}
    \label{tab:reward}
    \end{minipage} 
\end{table}

\begin{table*}[t!]
\centering
\begin{center}
\caption{\textbf{Combined Enhancement of Different Dimensions.} We combine the training data of related dimensions. 4 different combinations are included.}
\vspace{-2pt}
%\resizebox{0.98\linewidth}{!}{
\tablestyle{8pt}{1.2}
\begin{tabular}{l|c|c|c|c|c|l|c|c|c|c}
\toprule
\textbf{Aspect}&\multicolumn{5}{c|}{\textbf{Include Dimensions}}&\textbf{Aspect}&\multicolumn{4}{c}{\textbf{Include Dimensions}}\\
\midrule
\ding{172} Human& \textbf{\Centerstack{HAn}} & \textbf{\Centerstack{HCl}} & \textbf{\Centerstack{HIn}} & -&\textbf{Mean}&\ding{173} Physics&\textbf{\Centerstack{Mec}} & \textbf{\Centerstack{The}} & \textbf{\Centerstack{Mat}}&\textbf{Mean}\\
\midrule
Wan2.1&85.87 & 89.00 &72.33&-&82.40&Wan2.1&74.42&72.66&69.64&72.24\\
GigaVideo-1&\textbf{90.18}&\textbf{95.10}&\textbf{80.67}&-&\textbf{88.65}&GigaVideo-1&\textbf{76.30}&\textbf{73.19}&72.32&\textbf{73.94}\\
Combined&88.21&89.90&76.67&-&84.93&Combined&75.18&69.57&\textbf{75.89}&73.55\\
\midrule
\ding{174} Dynamics&\textbf{\Centerstack{DSR}} & \textbf{MOU} & \textbf{MoR} & \textbf{\Centerstack{DAt}}&\textbf{Mean}&\ding{175} Exisitence&\textbf{\Centerstack{Com}} &\textbf{\Centerstack{IPr}} &-&\textbf{Mean}\\
\midrule
Wan2.1&22.22&29.97&40.80&46.15&34.79&Wan2.1&44.23&63.26&-&53.75\\
GigaVideo-1&26.57&33.67&\textbf{52.87}&49.08&40.55&GigaVideo-1&47.96&69.03&-&58.50\\
Combined&\textbf{26.87}&\textbf{37.50}&47.70&\textbf{55.68}&\textbf{41.94}&   Combined&\textbf{50.74}&\textbf{70.80}&-&\textbf{60.77}\\
\bottomrule
\end{tabular}
%}
\vspace{-17pt}
\label{tab:combination}
\end{center}
\end{table*}

\subsection{Combined Enhancement of Different dimensions}
Beyond evaluating the effectiveness of our pipeline on individual dimensions, we further explore interactions and joint optimization across multiple dimensions, as shown in Tab.~\ref{tab:combination}. We begin by grouping semantically related dimensions into 4 broader categories based on conceptual overlap: Human, Dynamic, Physics, and Existence. For each group, we use multi-dimensional reward scores to jointly train the model across all dimensions within that group.

As shown in Tab.~\ref{tab:combination}, the results vary across groups. For the Human and Physics categories, joint training does not outperform single-dimension tuning on most metrics. In contrast, for Dynamic and Existence, joint optimization leads to consistent gains across the majority of constituent dimensions. We hypothesize that the limited gains in the Human and Physics categories arise from conflicting objectives across their dimensions. While semantically related, these dimensions often target diverse aspects that may interfere during joint training. In contrast, dimensions within the Dynamic and Existence groups tend to have more consistent goals, allowing for more effective joint optimization. These findings highlight the need for clearer dimension definitions and more deliberate prompt design to ensure aligned supervision across multiple evaluation axes.

\section{Conclusion}
We present GigaVideo-1, a lightweight fine-tuning framework that significantly enhances text-to-video generation.
By leveraging our prompt-driven data engine, our method effectively improves video quality by targeting underperforming aspects such as physical consistency and temporal logic. This approach ensures more focused and efficient model adaptation, without the need for large-scale data collection or manual annotations.
Secondly, the proposed reward-guided optimization utilizes automatic feedback to further refine model performance. Adaptively adjusting the training process, it leads to consistent and realistic outputs.
These two innovations enable GigaVideo-1 to achieve substantial improvements in video generation quality while minimizing computational cost and data requirements, offering a scalable and practical solution for advancing video generation models.

\newpage
\normalem
\bibliographystyle{unsrt}
\bibliography{ref}

\newpage
\appendix
\onecolumn
\setcounter{table}{4}
\setcounter{figure}{3}
\section{Experiments Details}
\noindent\textbf{Models and Settings.} We specify key hyperparameters and architectural choices here for reproducibility. We use the Qwen2.5-7B-Instruct~\cite{yang2024qwen2} model as our LLM for prompt augmentation and question generation for VQA, while LLaVA-Video-7B-Qwen2~\cite{zhang2024video} serves as the reward model to score synthesized videos. As the baseline T2V generation model, we choose Wan2.1-T2V-1.3B~\cite{wan2025wan}, applying full-parameter fine-tuning on its transformer to improve its expressiveness. We set the target resolution of video generation to 832×480 and generate 81 frames per video. During training, the batch size is set to 4 and the $\lambda_{kl}$ is set to 0.3 empirically. The learning rate is $1e-6$, with a single training epoch. This lightweight training schedule is designed to avoid overfitting and to preserve the general-purpose generation capabilities of the pretrained model. Details of the 6 state-of-the-art T2V models in Tab.1 can be found in Tab.~\ref{tab:model_info}.

\begin{table}[h]
\centering
\begin{center}
\caption{\textbf{Information on Evaluated Models.}}
\resizebox{0.75\linewidth}{!}{
\renewcommand\arraystretch{1.5}
\begin{tabular}{c|cccc}
\toprule
 \textbf{\Centerstack{Model Name}} &
\textbf{\Centerstack{Video Length}} 
& \textbf{\Centerstack{Per-Frame Resolution}} & 
\textbf{\Centerstack{Frame Rate (FPS)}} &
\textbf{\Centerstack{Frame Count}} 
\\ \midrule% \Xhline{1pt}
HunyuanVideo~\cite{kong2024hunyuanvideo}    & 5.3s & 720$\times$1280 & 24 &161 \\
CogVideoX-1.5~\cite{yang2024cogvideox}    & 10.1s & 768$\times$1360 & 16 &129 \\ 
Sora-480p~\cite{openaisora2024}    & 5.0s & 480$\times$854 & 30 &150 \\
Kling 1.6~\cite{kuaishou2024kling}    & 10.0s & 720$\times$1280 & 24 &241 \\ 
Wan2.1~\cite{wan2025wan}      &5.0s&480$\times$832&16 &81\\
GigaVideo-1      &5.0s&480$\times$832&16&81 \\
\bottomrule
\end{tabular}
}
\vspace{-8pt}
\label{tab:model_info}
\end{center}
\end{table}

\noindent\textbf{Datasets and Implementation.} The training data comprises two primary sources, as described in Sec. 3.3. First, the real-world dataset includes approximately 3.5k high-quality video samples selected from the Koala~\cite{wang2024koala} dataset. Each sample is paired with an accurate and detailed caption, along with a corresponding synthetic video generated using that caption. To align with the duration of training data used in Wan2.1, we retain only Koala videos shorter than 5 seconds. Further filtering is conducted using the Video Training Suitability Score provided by Koala, which is computed via the Training Suitability Assessment Network proposed in its paper. Second, the synthetic dataset consists of approximately 9.5k videos generated from domain-specific prompts produced by our prompt-driven data engine. These prompts are designed to emphasize specific generative aspects and support targeted fine-tuning across diverse quality dimensions.

\section{Prompt Generation}
We start by identifying specific failure modes in the pre-trained model (e.g., unrealistic motion, implausible interactions) and selecting a few representative prompt examples that expose these issues. Each weak dimension is defined with a short description and a handful of seed phrases. These are then fed into a large language model using few-shot prompting to generate a diverse set of base prompts that are concise, visually grounded, and explicitly focused on the target concept (e.g., “Camera zoom in, Disneyland.”). 

\begin{figure}[h]
    \centering
    \includegraphics[width=\linewidth]{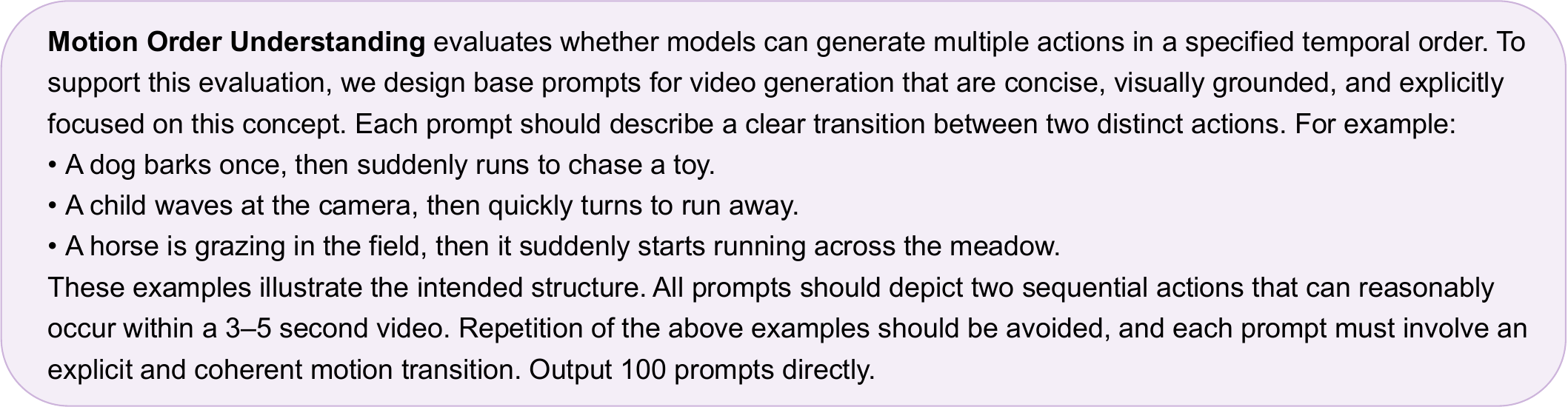} % Replace with your actual pipeline figure
    %\fbox{\rule{0pt}{2in} \rule{0.9\linewidth}{0pt}} % Placeholder box
    % \vspace{-1em}
    \caption{LLM prompts we use in the first stage for analogy and sentence making, targeting the Motion Order Understanding dimension as an example.}
    \label{fig:prompt} % Renamed label
    \vspace{-5pt}
\end{figure}

Specifically, we provide an example prompt targeting the Motion Order Understanding dimension to illustrate the generation process of the base prompts, as shown in the Fig.~\ref{fig:prompt}. For the second augmentation stage, in order to maintain consistency with the data distribution of the pre-trained model, we adopt the same prompt augmentation template provided by Wan2.1.

\section{Reward Models for Scoring}
Given that MLLMs are already capable of accurately understanding video scenes and events in most cases, we use them to score the performance of synthesized videos across the majority of evaluation dimensions. However, current MLLMs exhibit limitations in fine-grained visual perception, making their scores less reliable for a few specific dimensions that require such precision. For these cases, we employ specialized pre-trained visual models to provide more accurate assessments. For instance, in the instance preservation dimension, we use YOLO-World for object detection and counting; in the camera motion dimension, we use Cotracker2 for point tracking. As more capable MLLMs or scoring models become available in the future, the scoring module of GigaVideo-1 can be directly upgraded, further enhancing the reliability of reward-guided optimization.

\section{User Study}
We additionally conduct a user study to assess the quality preference of videos generated by Wan2.1 and GigaVideo-1 across the aforementioned evaluation dimensions. For each evaluation dimension, we randomly select five prompts and generate paired video samples using both Wan2.1 and GigaVideo-1 based on the same prompt. This results in pairwise comparison options, where the source model of each video is hidden and the order of presentation is randomized to prevent bias. 

\begin{table*}[h]
\centering
\begin{center}
\caption{\textbf{User Study of Different Dimensions.} This table presents evaluation results for our baseline Wan2.1 and GigaVideo-1 across 17 VBench-2.0 dimensions. A higher score indicates higher user preference in the corresponding dimension.
}
\vspace{-2pt}
\resizebox{0.98\linewidth}{!}{
\renewcommand\arraystretch{1.5}
\begin{tabular}{c|c|c|c|c|c|c|c|c|c}
\toprule
\multirow{2}{*}{\textbf{Models}}   & \textbf{Human} & \textbf{Human} & 
\textbf{Human} & \multirow{2}{*}{\textbf{Composition}} & \multirow{2}{*}{\textbf{Mechanics}} & \multirow{2}{*}{\textbf{Material}} & \multirow{2}{*}{\textbf{Thermotics}} & \textbf{Multi-view}&\textbf{Dynamic Spatial}\\
& \textbf{Anatomy} & \textbf{Clothes} & \textbf{Identity} & & & & & \textbf{Consistency}&\textbf{Relationship}\\
\midrule% \Xhline{1pt}
GigaVideo-1>Wan2.1~\cite{wan2025wan}&79.0\%&82.5\%&90.0\%&92.5\%&87.5\%&85.0\%&87.5\%&92.5\%&97.5\%\\

\toprule
\multirow{2}{*}{\textbf{Models}} &   \textbf{Dynamic} & \textbf{Motion Order} & \textbf{Human} & \textbf{Complex} & \textbf{Complex} & \textbf{Camera} & \textbf{Motion} & \textbf{Instance}& \multirow{2}{*}{\textbf{Mean}}\\
&   \textbf{Attribute} & \textbf{Understanding} & \textbf{Interaction} & \textbf{Landscape} & \textbf{Plot} & \textbf{Motion} & \textbf{Rationality} & \textbf{Preservation}&\\
% \Xhline{1pt}
\midrule
GigaVideo-1>Wan2.1~\cite{wan2025wan}&85.0\%&90.0\%&82.5\%&95.0\%&100.0\%&92.5\%&80.0\%&95.0\%&88.8\%\\
\bottomrule
\end{tabular}
}
\vspace{-4mm}
\label{tab:userstudy}
\end{center}
\end{table*}

Annotators are shown the name of the target dimension and asked to choose which video they prefer with respect to that specific aspect. Each dimension is evaluated by 10 annotators, and we compute the user preference score of GigaVideo-1 over Wan2.1 as the percentage of comparisons where GigaVideo-1 is preferred. As shown in Tab.~\ref{tab:userstudy}, GigaVideo-1 achieves preference scores above 50\% on the majority of dimensions, demonstrating clear improvements in perceptual quality brought by our reward-guided optimization.

\section{All Dimensions Evaluation}
Beyond evaluating the effectiveness of our pipeline on individual dimensions and some related dimensions, we further explore the joint optimization across all 17 dimensions defined by VBench-2.0, as shown in Tab~\ref{tab:combined2}.
\begin{table*}[h]
\large
\centering
\begin{center}
\caption{\textbf{Combined Enhancement of All 17 Dimensions of VBench-2.0.} This table presents evaluation results of our baseline Wan2.1 and the combined enhancement version of GigaVideo-1 across 17 VBench-2.0 dimensions. A higher score indicates better performance of this dimension.
}
\vspace{-2pt}
\resizebox{0.98\linewidth}{!}{
\renewcommand\arraystretch{1.5}
\begin{tabular}{c|c|c|c|c|c|c|c|c|c|c|c|c|c|c|c|c|c|c}
\toprule
\textbf{Models}   & \textbf{\Centerstack{HAn}} & \textbf{\Centerstack{HCl}} & 
\textbf{\Centerstack{HId}} & \textbf{\Centerstack{HIn}} & \textbf{\Centerstack{DSR}} & \textbf{MOU} & \textbf{MoR} & \textbf{\Centerstack{DAt}} & \textbf{\Centerstack{Mec}} &  \textbf{\Centerstack{The}} & \textbf{\Centerstack{Mat}} & \textbf{\Centerstack{Com}} & \textbf{\Centerstack{IPr}} & \textbf{CMt} & \textbf{\Centerstack{MVC}} & \textbf{\Centerstack{CoL}} & \textbf{\Centerstack{CoP}} & \textbf{Mean} \\ 
\midrule
%\multicolumn{18}{c}{Single Dimension Enhancement} \\
%\midrule
Wan 2.1~\cite{wan2025wan} & 85.87 & 89.00 & 67.02&72.33&22.22&29.97&40.80&46.15&\textbf{74.42}&\textbf{72.66}&69.64&44.23&63.26&\textbf{36.11}&\textbf{44.60}&17.11&\textbf{10.69}&52.12 \\
\textbf{All} &\textbf{89.88}&\textbf{91.26}&\textbf{67.33}&\textbf{73.33}&\textbf{29.47}&\textbf{37.71}&\textbf{51.15}&\textbf{47.99}&70.53&65.94&\textbf{75.22}&\textbf{49.78}&\textbf{71.41}&35.49&33.55&\textbf{22.22}&10.36&\textbf{54.27} \\
% \Xhline{1pt}
\bottomrule
\end{tabular}
}
\vspace{-4mm}
\label{tab:combined2}
\end{center}
\end{table*}

Joint optimization generally yields good performance across most dimensions compared to our baseline Wan2.1. However, certain dimensions exhibit diminished gains. We hypothesize that this may be due to conflicting objective signals arising from inconsistent definitions or incompatible prompt formulations across different dimensions. Addressing this issue may require more principled dimension definitions and prompt engineering strategies to ensure coherent supervision across multiple axes of evaluation.

\section{Additional Analysis of Reward Strategies}
To further understand the role of reward signal acquisition and integration, we compare several optimization strategies. In Tab.4, 1).``Reweight-offline'' uses offline scores computed from synthetic videos prior to training. 2).``Reweight-online'' predicts reward scores during training based on intermediate denoised frames. 3).``Backprop-online'' directly incorporates reward scores into the loss function itself, allowing gradient backpropagation through the reward signal. 

In Tab.4, reweighting with online scores yields slightly lower performance and incurs higher computational cost, likely due to the limited reliability of intermediate denoised frames used during scoring. Backpropagating through the reward signal improves accuracy marginally but further increases the training burden, making it less practical in resource-constrained settings. Among them, our offline reweighting achieves the best high accuracy with minimal training overhead. These results demonstrate that our reward-guided optimization framework offers a highly efficient and effective alternative to conventional fine-tuning.

\section{Broader Impact}
GigaVideo-1 presents a scalable and efficient pipeline for enhancing text-to-video generation, offering improved alignment with human preferences through automated reward modeling and dimension-specific fine-tuning. By significantly reducing the reliance on manual annotation and leveraging multimodal feedback, GigaVideo-1 lowers the barrier for high-quality video generation and has the potential to benefit a wide range of applications, including digital storytelling, educational content creation, and simulation prototyping. Several potential risks warrant attention. First, the reliance on pretrained foundation models means that GigaVideo-1 may inherit biases present in its underlying components, including in the scoring signals used for reward optimization. This could lead to unintended reinforcement of stereotypes or systematic errors if not properly audited. Second, this increased accessibility may bring the risk of misuse. Like other generative video technologies, GigaVideo-1 could be exploited to produce misleading or manipulated content, such as fabricated news footage or synthetic evidence. To mitigate this, it is important to promote responsible deployment practices, encourage transparency in generated media, and support ongoing research into detection and watermarking tools. Lastly, while our method improves specific dimensions of generation quality, it may lead to trade-offs in aspects not explicitly optimized, highlighting the importance of balanced, multi-objective evaluation and ongoing community benchmarking efforts. We encourage future research to explore more robust alignment signals, diversified data sources, and transparent evaluation protocols to ensure that methods like GigaVideo-1 contribute to the responsible and inclusive advancement of video generation technologies.

\end{document}